\documentclass[letterpaper, 10 pt, conference]{ieeeconf}  
\IEEEoverridecommandlockouts                              

\usepackage{xcolor}
\usepackage{amssymb}
\usepackage{graphicx}
\usepackage{mathtools}
\usepackage{amsmath}
\usepackage{gensymb}
\usepackage{float}
\usepackage{multirow}
\usepackage{makecell}
\usepackage{mathtools}
\usepackage{hyperref}
\hypersetup{
    colorlinks=true,
    linkcolor=blue,
    filecolor=magenta,      
    urlcolor=blue,
    pdftitle={Overleaf Example},
    pdfpagemode=FullScreen,
    }

\usepackage{tabularx}
    \newcolumntype{L}{>{\raggedright\arraybackslash}X}
\usepackage[utf8]{inputenc}
\usepackage[english]{babel}

\usepackage{amsthm}

\usepackage{subcaption}

\newcommand{\no}{\noindent}

\newtheorem{prop}{Proposition}[section]
\newtheorem{probform}{Formulation}[section]

\newcommand\Tstrut{\rule{0pt}{2.6ex}}         
\title{\LARGE \bf VERN: Vegetation-aware Robot Navigation in Dense Unstructured Outdoor Environments
}

\author{Adarsh Jagan Sathyamoorthy$^1$, Kasun Weerakoon$^1$, Tianrui Guan,$^1$ Mason Russell$^2$, \\ Damon Conover$^2$, Jason Pusey$^2$ and Dinesh Manocha$^1$ 
\thanks{This work has been conducted as part of the ArtIAMAS cooperative agreement (https://artiamas.umd.edu) and is being partially funded by Army Research Laboratory Grant No. W911NF2120076.}
}

\begin{document}

\maketitle
\thispagestyle{empty}
\pagestyle{empty}

\footnotetext[1]{Authors are with the University of Maryland, College Park.}

\footnotetext[2]{Authors are with the DEVCOM Army Research Laboratory, Aberdeen Proving Ground, Maryland, USA.}

\begin{abstract}
We propose a novel method for autonomous legged robot navigation in densely vegetated environments with a variety of pliable/traversable and non-pliable/untraversable vegetation. We present a novel few-shot learning classifier that can be trained on a few hundred RGB images to differentiate flora that can be navigated through, from the ones that must be circumvented. Using the vegetation classification and 2D lidar scans, our method constructs a vegetation-aware traversability cost map that accurately represents the pliable and non-pliable obstacles with lower, and higher traversability costs, respectively. Our cost map construction accounts for misclassifications of the vegetation and further lowers the risk of collisions, freezing and entrapment in vegetation during navigation. Furthermore, we propose holonomic recovery behaviors for the robot for scenarios where it freezes, or gets physically entrapped in dense, pliable vegetation. We demonstrate our method on a Boston Dynamics Spot robot in real-world unstructured environments with sparse and dense tall grass, bushes, trees, etc. We observe an increase of 25-90\% in success rates, 10-90\% decrease in freezing rate, and up to 65\% decrease in the false positive rate compared to existing methods.
\end{abstract}

\section{Introduction} \label{sec:Intro}

In recent times, mobile robots have been used for many outdoor applications in agriculture \cite{harvesting-robots,milioto2018real,borges2022survey} (automatic seeding, harvesting, and measuring plant and soil health), gardening \cite{trimbot}, forest exploration, search and rescue \cite{borges2022survey}, etc. Operating in such environments entails navigating in the presence of vegetation with varying height, density, and rigidity. 

In such dense and unstructured vegetation, the robot may not always find free space to circumvent the flora. This causes the robot to freeze \cite{frozone}; a phenomenon where its planner cannot compute any collision-free velocity to reach its goal. The robot either halts or starts oscillating indefinitely, leading to collisions and not progressing to its goal. Additionally, a small wheeled or legged robot could get physically entrapped in vegetation when its wheels or legs get intertwined in dried tall grass, bushes, etc. In such cases, the robot's dynamics determines whether it can autonomously recover itself \cite{vision-aided-quadruped}. 

To effectively navigate such environments, the robot must assess the traversability of the various flora around it. Firstly, it must differentiate flora based on their \textit{pliability} \cite{char-traversal-pliable-veg}. We define a plant's pliability as its degree of flexibility or ability to bend such that a robot can navigate through it. For instance, robots can traverse \textit{through} tall grass (high pliability) whereas trees are non-pliable and should be avoided. Secondly, the robot must detect the height and density of pliable vegetation since they affect the resistance offered to the robot's motion. 


A major challenge in differentiating pliable/traversable from non-pliable/untraversable flora around the robot stems from sensors (e.g., laser scans, point clouds, and ultrasound) detecting all plants as solid obstacles \cite{badgr}. Furthermore, vegetation such as tall grass scatters laser beams from lidars and leads to poor characterization of their shape and structure \cite{multiwavelength-lidar}. In RGB images, the shape and structure of various plants are accurately represented. However, there is a lack of research on differentiating flora based on their pliability from images. 

\begin{figure}[t]
    \centering
    \includegraphics[width=\columnwidth,height=5.5cm]{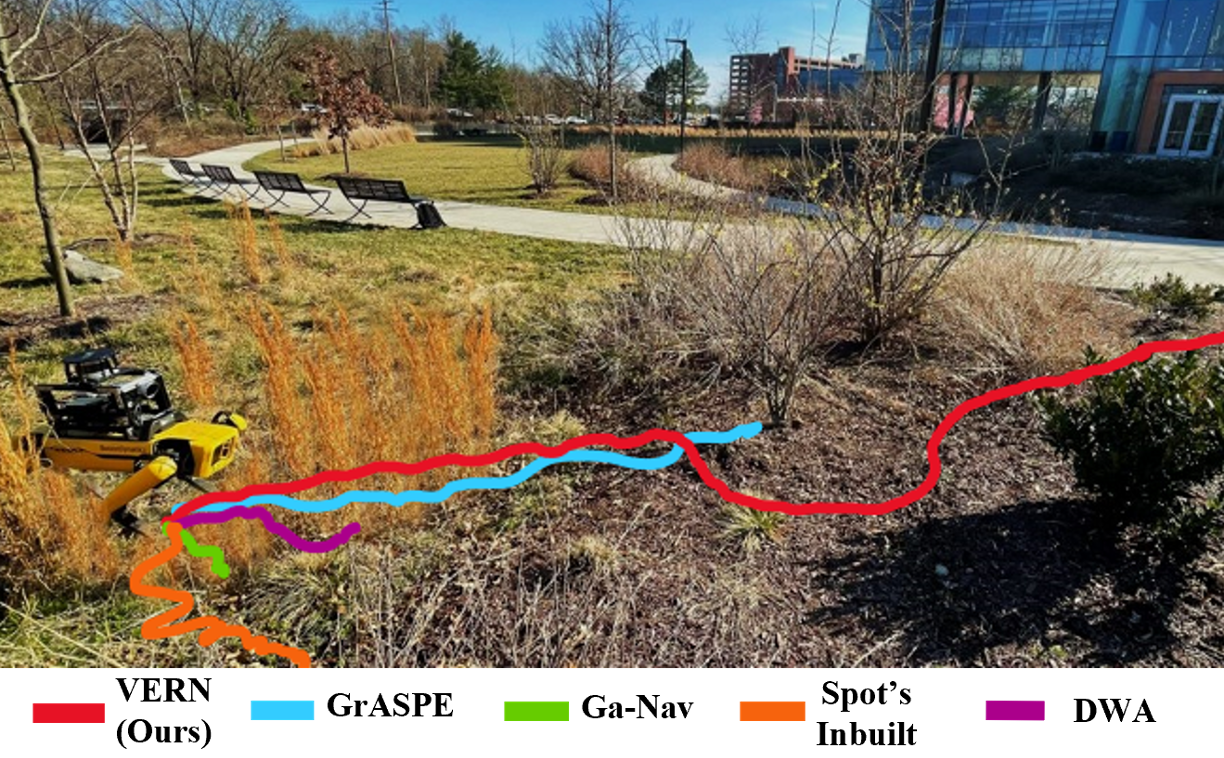}
    \caption{\small{Comparison of VERN with other methods navigating a Spot robot through a complex environment (scenario 3) with traversable (e.g. tall grass), and untraversable (e.g. tree, bush) vegetation. In this trial, we observe that only VERN successfully reaches its goal due to its vegetation classification, height estimation, and novel cost map clearing scheme. Other methods either collide (GrASPE \cite{graspe}, Spot's in-built autonomy), or freeze (DWA \cite{fox1997dwa}, GA-Nav \cite{ganav}) due to the density of vegetation.}}
    \label{fig:cover_image}
    \vspace{-15pt}
\end{figure}

There are several works in computer vision to detect and segment simple vegetation such as trees, and short grass \cite{milioto2018real,ganav,semantic-mapping-auto-off-road-nav}. However, such works require extensive datasets with intricate annotations to train, requiring significant human effort. Additionally, segmenting traversable versus untraversable regions in dense vegetation is still an unsolved problem \cite{offroad-mobility}.

Navigation methods for outdoor domains have dealt with flora such as tall grass \cite{badgr}, trees, etc. in isolation. Such methods do not consider scenarios where plants of all pliability/traversability are in close proximity. Therefore, such methods cause the robot to collide or get entrapped, and do not have autonomous mechanisms to recover in such environments. Navigation methods based on end-to-end learning \cite{badgr,model-error-katyal} require real-world negative experiences such as collisions with non-pliable obstacles for training. This would be impractical for highly unstructured regions where rates of collision and entrapment are high. 

\textbf{Main Contributions:} To address these limitations and navigate in densely vegetated environments, we present VERN (\textbf{VE}getation-aware \textbf{R}obot \textbf{N}avigation) using multi-modal perception sensors. The novel components of our work include:



\begin{itemize}
    \item A novel classifier to differentiate vegetation of various pliability/traversability from RGB images. We propose a novel few-shot learning approach and a siamese network framework to train our classifier with only a few hundred images of different kinds of vegetation (tall grass, bushes/shrubs, and trees) with minimal human labeling. This is a significant decrease in the required dataset size and human effort compared to existing methods for outdoor terrain perception. 
    

    \item A novel method to compute a \textit{vegetation-aware} traversability cost map by accounting for estimated vegetation height, pliability classification, and the classification confidence. Our method accounts for misclassifications in the vegetation classifier and leads to a more accurate representation of the traversable vegetation in the robot's surroundings in the cost map. This leads to an improvement of 25-90\% in terms of success rate and a decrease of up to 65\% false positive rate.


    \item A local planner that produces cautious navigation behaviors when in highly unstructured vegetation or regions with low confidence pliability classification. Additionally, the planner initiates novel holonomic behaviors to recover the robot if it freezes or gets physically entrapped in dense vegetation. This improves the success rate by up to 50\%, and the freezing rate by 75\%.
\end{itemize}

\section{Related Work}

In this section, we discuss previous works in outdoor navigation and perception of unstructured vegetation.

\subsection{Perception in Dense Vegetation}
Early works on vegetation detection were typically chlorophyll detectors \cite{Bradley-2004-8857,nguyen-1,double-check-passable}, or basic classification models \cite{auto-terrain-characterization-modeling}.  Nguyen et al. \cite{nguyen-1} proposed a sensor setup to detect near-IR reflectance and used it to define a novel vegetation index. \cite{double-check-passable} extended this detection by using an air compressor device to create strong winds and estimating the levels of resistance to robot motion using the movement of vegetation. However, it needed the robot to be static. Multiple sensor configurations have also been explored for obstacle detection within vegetation \cite{sensing-tech-obstacle-detect} of which thermal camera and RADAR stand out as effective modalities.

Modeling the frictional/lumped-drag characteristics of vegetation \cite{momentum-traversal-mobility-challenges,char-traversal-pliable-veg,thesis} as a measure of the resistance offered to motion, or modeling plant stems using their rotational stiffness and damping characteristics \cite{modeling-traversal-pliable-materials} has also been proposed. However, the method does not use visual feedback, requiring the robot to drive through vegetation first to gauge its pliability.

Wurm et al. \cite{veg-laser-data-structured-outdoor,wurm2009improving} presented methods to detect short grass from lidar scans based on its reflective properties on near IR light. Astolfi et al. \cite{vineyard} demonstrated accurate SLAM and navigation through sensor fusion in a vineyard. However, such methods operated in highly structured environments. 

There are several works that utilize semantic segmentation to understand terrain traversability \cite{ganav,semantic-mapping-auto-off-road-nav,terrain-semantics-multi-legged}. \cite{terrain-semantics-multi-legged} applied semantic segmentation classifier trained on RGB images to oct-tree maps obtained from RGB-D point cloud. Maturana et al. \cite{semantic-mapping-auto-off-road-nav} took a similar approach by training a segmenter and augmenting a 2.5D grid map with it to distinguish tall grass from regular obstacles. However, these existing methods are not suitable for perception in dense, unstructured vegetation.

\subsection{Navigation in Unstructured Vegetation}
Although there are many works on outdoor, off-road navigation to handle slopes \cite{terp} and different terrain types \cite{terrapn}, there have been only a few methods for detecting and navigating through vegetation. \cite{Overbye-1} proposed using a terrain gradient map along with the A* algorithm to navigate a large wheeled robot and demonstrated moving over tall grass and short bushes. \cite{Overbye-2} extended this by adding a segmentation layer to the map to detect soft obstacles. However, these methods mostly operate in structured, isolated vegetation and may not work well in unstructured scenarios.


There are several works that have addressed navigating through pliable vegetation such as tall grass \cite{badgr,model-error-katyal}. Kahn et al. \cite{badgr} demonstrated a model that learns from a robot's real-world experiences such as collisions, bumpiness, and its position to navigate outdoor environments. The robot learned from RGB images and associated experiences (labels) to consider tall grass as traversable. However, it does not account for the presence of non-traversable bushes or trees alongside traversable vegetation. Polevoy et al. \cite{model-error-katyal} proposed a model that regresses the difference between the robot's dynamics model and its actual realized trajectory in unstructured vegetation. This acts as a measure of the surrounding vegetation and terrain's traversability. However, both these methods require ``negative" examples such as collisions during training, which may be impractical or dangerous to collect in the real world. 


With the advent of legged robots, several recent works have been focused on developing robust controllers for locomotion purely using proprioception \cite{quadruped-locomotion-challenging-terrain} or fusing it with exteroceptive perception such as elevation maps \cite{robust-perceptive-locomotion-quad}.  A few navigation works focus on estimating the underlying support surface that is hidden by vegetation \cite{support-surface-legged-robot} by fusing haptic feedback from the robot, depth images, and detecting the height of the vegetation. Our method is complementary to these existing works.

\section{Background}
\begin{figure}[t]
      \centering
      \includegraphics[width=\columnwidth,height=4.0cm]{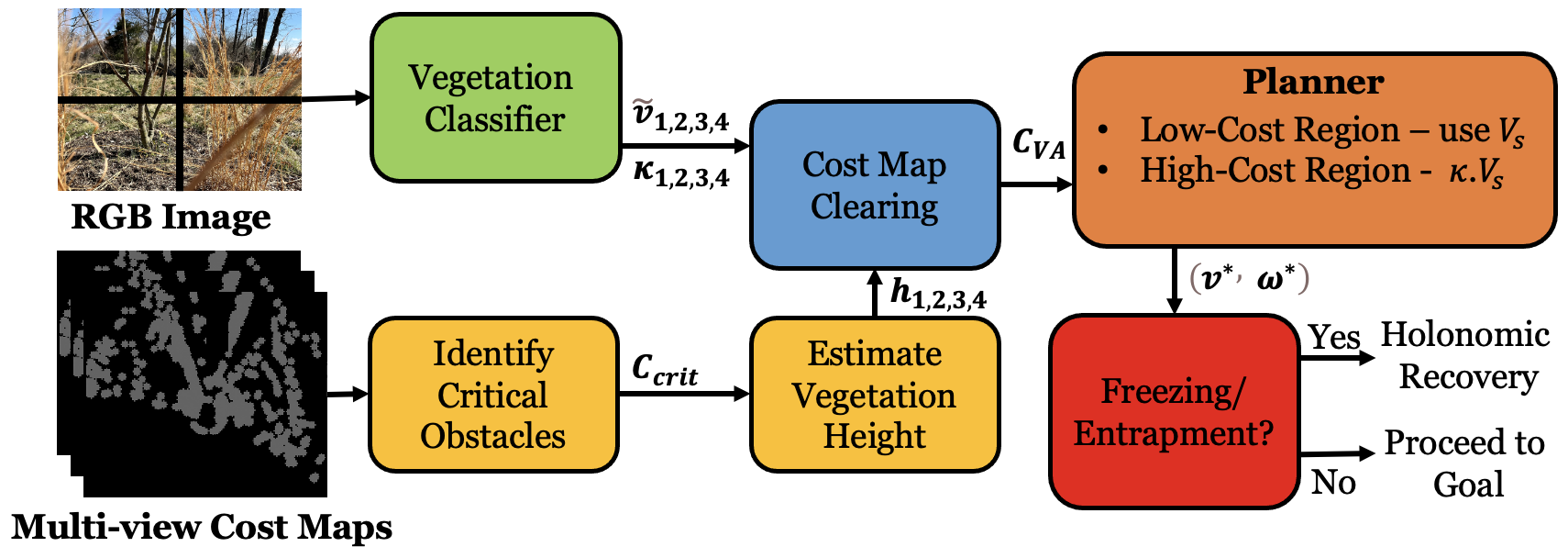}
      \caption {\small{VERN's overall system architecture for vegetation classification to estimate pliability/traversability. Our model uses classification, its confidence, and vegetation height to create a vegetation-aware cost map $C_{VA}$ used for planning. In high traversability cost regions, our planner executes cautious behaviors by limiting the robot's velocity space. If the robot freezes or gets entrapped, it executes holonomic recovery behaviors.}  }
      \label{fig:sys-arch}
      \vspace{-10pt}
\end{figure}

In this section, we provide our problem formulation, briefly explain siamese networks used for vegetation classification, and the Dynamic Window Approach (DWA) \cite{fox1997dwa} used as a basis for our navigation. Our overall system architecture is shown in Fig. \ref{fig:sys-arch}.

\subsection{Setup and Conventions}
We assume a legged robot setup with holonomic dynamics equipped with a 3D lidar (that provides point clouds and 2D laser scans) and an RGB camera. Certain quantities are measured relative to a global coordinate frame (referred to as odom hereafter) with the robot's starting location as its origin. Some quantities are relative to a cost map grid moving with and centered at the robot. The quantities can be transformed between these frames using transformation matrices $T^O_C$ or $T^C_O$. The reference frame for each quantity is indicated in the superscript ($O$ or $C$) of its symbol. We denote vectors in bold, and use symbols such as i, j, row, col as indices. Apart from its pose w.r.t odom, the robot does not have access to any global information.


\subsection{Problem Domain}
We consider navigation in environments containing three kinds of vegetation together: 1. Tall grass, 2. Bushes/shrubs, and 3. Trees. We make the following assumptions about the three types. 

\subsubsection{Properties of Vegetation}
Tall grass is pliable (therefore traversable), with height $> 0.3m$ (could be taller than the height the robot's sensors are mounted at), and could be of varying density in the environment. Bushes/shrubs (height $0.1 - 0.5m$), and trees (height $> 2m$) are non-pliable/untraversable and must be circumvented. 


\subsubsection{Adverse Phenomena} \label{sec:adverse-phen}
Navigating through such vegetation, the robot could encounter three adverse phenomena: 1. Freezing, 2. Physical entrapment in vegetation, and 3. Collisions. The conditions for these phenomena are, 
\begin{enumerate}
    \item Freezing:  $V_r = \varnothing$,
    
    \item Entrapment: $(v^*, \omega^*) \ne (0, 0)$ but $\Delta x^{O}_{rob} = \Delta y^{O}_{rob} = \Delta \theta^{O}_{rob} = 0$,

    \item Collisions: $dist(\text{robot}, \text{\{bushes/trees\}}) = 0$.
\end{enumerate}

\no Here, $V_r$ refers to the robot's feasible, collision-free velocity space. $(v^*, \omega^*)$ is the velocity commanded by the robot's planner, and $\Delta x^{O}_{rob}, \Delta y^{O}_{rob}, \Delta \theta^{O}_{rob}$ refer to the robot's change in position and yaw orientation relative to the global odom frame. 
Freezing occurs in densely vegetated environments if the planner cannot find any collision-free velocity to reach its goal. To avoid this, the robot must accurately identify pliable vegetation to navigate through. The robot could also get physically entrapped in dense vegetation if its legs are caught in grass, small branches of bushes, etc. Finally, the robot could collide if its planner misclassifies a non-pliable obstacle as pliable. 

\no With these definitions, VERN's problem formulation can be stated as follows:
\begin{probform}
To compute a robot velocity $(v^*, \omega^*)$ such that $traj^O(v^*, \omega^*) \in Free^O \, \cup \, Grass^O$, $traj^O(v^*, \omega^*) \notin Tree^O \, \cup \, Bush^O$, assuming that $Tree^O \, \cap \, Grass^O = Tree^O \, \cap \, Bush^O = Bush^O \, \cap \, Grass^O = \varnothing$. 
\end{probform}

\no Here, $Tree^O, Bush^O, Grass^O$ represent the sets of locations containing trees, bushes, and grass, respectively that were detected by the robot w.r.t the odom frame. $Free^O$ represents free space locations. $traj^O()$ returns the forward-simulated trajectory of a $(v, \omega)$ pair relative to the odom frame. We assume that the robot can balance and walk on all the terrains we consider. We also assume that the robot never encounters a scene with no free space and only non-pliable vegetation (an unresolvable scenario). 



\subsection{Siamese Networks}
Siamese networks \cite{SiameseNN} are a class of neural networks used for one-shot or few-shot learning (learning with very few samples). They use two sub-networks with identical structures, parameters, and weights, and are connected at the end using an energy function (e.g., L2 norm). They accept two distinct images as inputs for the two sub-networks and are trained to map similar images close to each other in a low-dimensional latent space. Dissimilar images have latent representations that are mapped far away from each other. 



\subsection{Dynamic Window Approach} \label{sec:dwa}
Our planner adapts the Dynamic Window Approach (DWA)~\cite{fox1997dwa} to perform navigation. We represent the robot's actions as linear and angular velocity pairs $(v,\omega)$. Let $V_s = [[0, v_{max}], [-\omega_{max}, \omega_{max}]]$ be the space of all the possible robot velocities based on the maximum velocity limits. DWA considers two constraints to obtain dynamically feasible and collision-free velocities: (1) The dynamic window $V_d$ contains the reachable velocities during the next $\Delta t$ time interval based on acceleration constraints;  (2) The admissible velocity space $V_a$ includes the collision-free velocities. The resulting velocity space $V_r = V_s \cap V_d \cap V_a$ is utilized to calculate the optimal velocity pair $(v^*,\omega^*)$ by minimizing the objective function below,
\vspace{-4pt}
\begin{equation}
\label{eq:dwa_obj_func}
    Q(v,\omega) = \sigma\big(\gamma_1 . head(.) + \gamma_2 . obs(.) + \gamma_3 . vel(.) \big),
\vspace{-4pt}
\end{equation}
where $head(.)$, $obs(.)$, and $vel(.)$ are the cost functions~\cite{fox1997dwa} to quantify a velocity pair's (($v, \omega$) omitted on RHS for readability) heading towards the goal, distance to the closest obstacle in the trajectory, and the forward velocity of the robot, respectively. $\sigma$ is a smoothing function and $\gamma_i, (i=1,2,3)$ are adjustable weights.
\section{VERN: Vegetation Classification}
\begin{figure}[t]
      \centering
      \includegraphics[width=7.5cm,height=4.5cm]{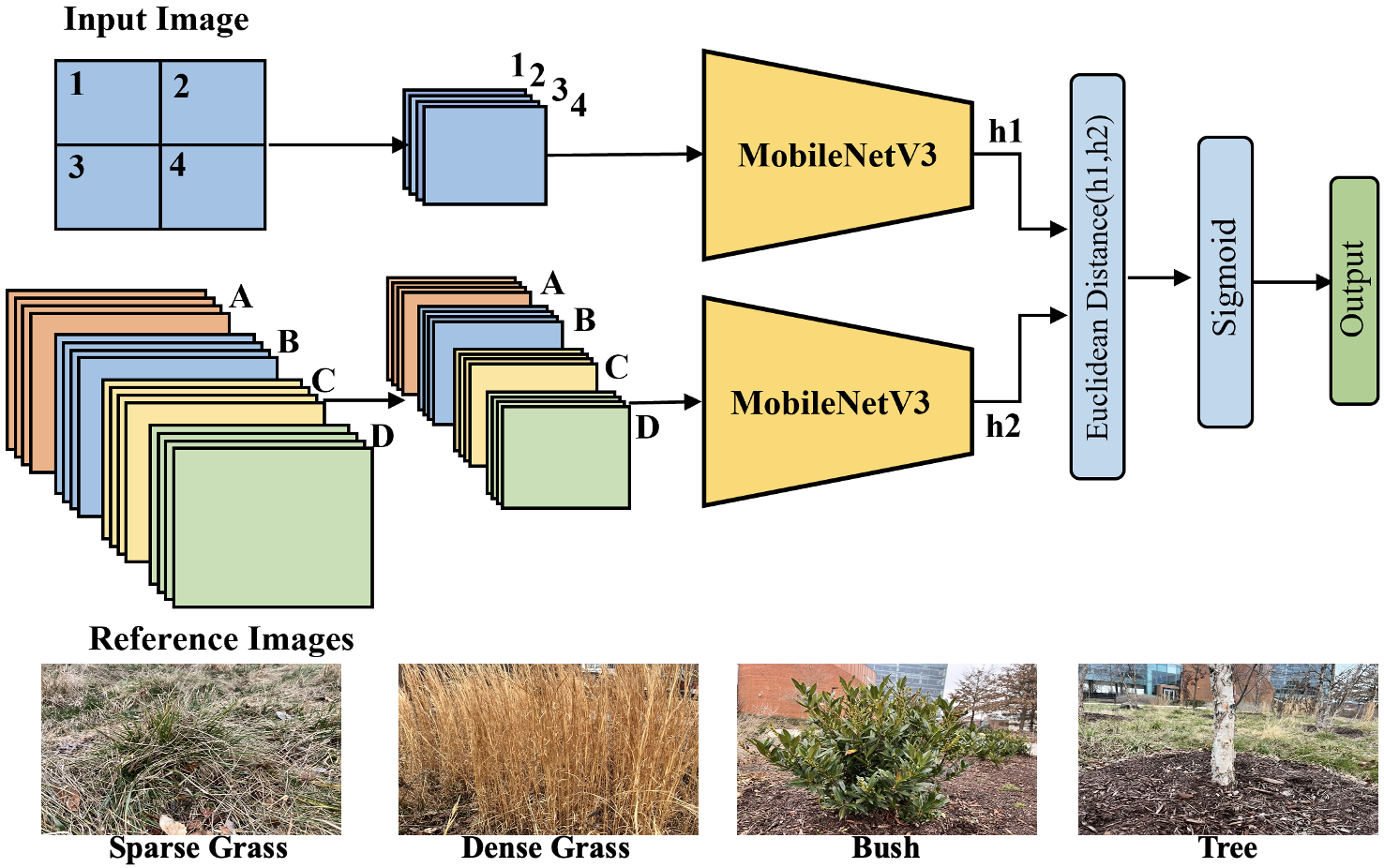}
      \caption {\small{VERN's classification network architecture. Quadrants of the real-time camera image are paired with several reference images for each class and fed into the two identical branches of our network. Example reference images for each class are shown at the bottom.}}
      \label{fig:network-arch}
      \vspace{-10pt}
\end{figure}

Our vegetation classifier uses an RGB image ($I^{RGB}_t$) obtained from a camera on the robot at a time $t$ as input. Although a plant's structure is well preserved in an image, using the entire image for classification is infeasible because the scene in $I^{RGB}_t$ typically contains two or more types of vegetation together. Therefore, we split $I^{RGB}_t$ into four quadrants $Q_1$ (top-left), $Q_2$ (top-right), $Q_3$ (bottom-left), $Q_4$ (bottom-right) as shown in Figs. \ref{fig:network-arch}, \ref{fig:costmap_comparisons}. A quadrant predominantly contains a single type of vegetation typically (see Fig. \ref{fig:costmap_comparisons}). 

We classify vegetation in the quadrants into four classes: 1. sparse grass, 2. dense grass, 3. bush, and 4. tree. We separate sparse and dense grass due to their visual dissimilarity. 


\subsection{Data Preparation}
To create the training dataset, images collected from different environments are split into quadrants and grouped together manually based on the vegetation type predominant in a quadrant. Next, pairs of images are created either from the same group or from different groups. The pair with similar images (same group) is automatically labeled 1, or 0 otherwise. The entire data preparation process takes about 30-45 minutes of manual effort. The obtained image pairs and the corresponding labels are passed into the classification network for training.

\subsection{Network Architecture}
Our vegetation classifier network consists of two identical feature extraction branches to identify the similarity between input image pairs. Our feature extraction branches are based on the MobileNetv3 \cite{mobilenetv3} backbone. We choose MobileNetv3 because it incorporates depth-wise separable convolutions (i.e., fewer parameters), leading to a comparatively lightweight and fast neural network. The outputs of the MobileNetv3 branches ($h_1$ and $h_2$) are one-dimensional latent feature vectors of the corresponding input images. The euclidean distance between the two feature vectors is calculated and passed through a $sigmoid$ activation layer to obtain the predictions.

We utilize the \textit{contrastive loss} function, which is capable of learning discriminative features to evaluate our model during training. Let $\hat{y}$ be the prediction output from the model. The contrastive loss function $J$ is:
\vspace{-5pt}
\begin{equation}
    J = \hat{y} \cdot d^2 + (1-\hat{y}) \cdot max(margin-d,0)^2,
\end{equation}

where $d$ is the euclidean distance between the feature vectors $h_1$ and $h_2$. $Margin$ is used to ensure that dissimilar image pairs are at least $margin$ distance apart.

\subsection{Network Outputs} \label{sec:net-outputs}
During run-time, the quadrants $Q_1, Q_2, Q_3,$ and $Q_4$ are each paired with several reference images of the four classes. The several reference images per class have different viewpoints and lighting conditions. They are fed into the classifier model $\mathcal{F}$ as a batch to obtain predictions as, 
\begin{equation}
    \mathcal{F}(Q_1, Q_2, Q_3, Q_4) = \Tilde{V}_{4 \times 4} |\,\, (\Tilde{V}_{ij} \in [0, 1]) \,\, i, j \in \{1, 2, 3, 4\} .
\end{equation}

\no $\Tilde{V}_{4 \times 4}$ is the output prediction matrix whose values $\Tilde{V}_{i,j}$ correspond to the \textit{least} euclidean distance of the quadrant $Q_i$ from any of the reference images for the $j^{th}$ class. The closer $\Tilde{V}_{i,j}$ is to 0, the higher the similarity between $Q_i$ and class $j$. 

We extract two outputs from $\Tilde{V}_{4 \times 4}$. Namely, for each quadrant $Q_i$: 1. the vegetation class that is most similar to it ($\Tilde{v}_i$), and 2. the corresponding similarity score ($d_{i}$):
\vspace{-5pt}
\begin{equation}
    \Tilde{v}_i = \underset{j}{\operatorname{argmin}} \Tilde{V}_{i,j}, \,\, \text{and} \,\, d_{i} = \Tilde{V}_{i,j}. 
\end{equation}

\no For readability, hereafter we denote $\Tilde{v}_i$ as belonging to the Pliable Vegetation (PV) set if $\Tilde{v}_i = 1$ or $2$ (sparse/dense grass), and to the Non-Pliable Vegetation (NPV) set if $\Tilde{v}_i = 3$ or $4$ (bush/tree). We define the confidence of classification $\kappa_i$ as,
\vspace{-7pt}
\begin{equation}
    \kappa_i = e^{-\alpha \cdot d_i}, 
\end{equation}
\no where $\alpha$ is a tunable parameter. We observe that $d_i \to 0 \implies \kappa_i \to 1$, and vice versa.


\section{VERN: Navigation in Dense Vegetation}
We use occupancy grids/cost maps generated using 2D lidar scans to detect obstacles around the robot. Our cost map can be defined as,

\vspace{-10pt}
\begin{equation}
    C_z(row, col) = \{p \,\, | \,\, p \in \{0, 100\}\}.
\end{equation}

\no Here, z is the height (along the robot's Z-axis) at which the 2D scan is recorded on a plane parallel to the ground. $p$ is a binary variable representing if an obstacle is present at $(row, col)$ of the cost map. In a densely vegetated environment, $C_{z}$ contains obstacles in all the locations where the lidar's scans have a finite proximity value. However, it does not account for the pliability of certain types of vegetation, which makes them passable and therefore not true obstacles. Therefore, we use the classification results, its confidence (Section \ref{sec:net-outputs}), and estimated vegetation height to augment our cost map prior to navigation.



\subsection{Multi-view Cost Maps}
To estimate the height of the environmental obstacles, we use three cost maps of equal dimensions $C_{low}, C_{mid},$ and $C_{high}$ corresponding to 2D scans from three different heights in 3D lidar's point clouds. The cost maps correspond to a height lower, equal to, and higher than the height at which the robot's 3D lidar is mounted, respectively. $C_{low}$ contains obstacles of all heights around the robot. $C_{mid}$ and $C_{high}$ contain taller obstacles such as tall grass, trees, walls, humans, etc. Using only three views instead of projecting the entire point cloud onto a 2D plane reduces the computation burden. It also helps identify truly tall obstacles and avoids misclassifying overhanging leaves from trees as tall obstacles.

We consider tall obstacles as \textit{critical obstacles} due to the high probability of them being solid obstacles such as trees and walls. To identify critical obstacles, we perform the element-wise sum operation as follows,
\vspace{-7pt}
\begin{gather}
    C_{crit} = C_{high} \bigoplus C_{mid} \bigoplus C_{low}.
\end{gather}

\no $C_{crit}(row, col) \in \{0, 100, 200, 300\}$, and regions with higher costs contain critical obstacles.


\subsection{Vegetation-aware Cost Map} \label{sec:costmap-clearing}
To combine $\Tilde{v}_i$, $\kappa_i$ and $C_{crit}$, we must correlate the regions viewed by quadrants $Q_{1,2,3,4}$ in $I^{RGB}_t$ with the corresponding regions in $C_{crit}$. 

\subsubsection{Homography}
To this end, we apply a homography transformation $H$ to project $I^{RGB}_t$ onto the cost map. Let $reg()$ denote a function that returns the real-world $(x, y)$ coordinates corresponding to a region of a cost map or image. We obtain the image quadrant relative to the cost map (see green/red rectangles in Fig. \ref{fig:costmap_comparisons} bottom) as,  
\vspace{-3pt}
\begin{equation}
    Q^C_i = \{ (row, col) | reg(C(row, col)) = reg(H(Q_i)) \}.
\end{equation}

\subsubsection{Cost Map Clearing}

We now use $\Tilde{v}_i$, $\kappa_i$, $C_{crit}$, and $Q^C_i$ to clear/modify the costs of the grids in $C_{low}$. We choose to modify and plan over $C_{low}$ since it detects obstacles of all heights. First, we calculate the normalized height measure (between 0 to $\pi/2$) of the obstacles in each quadrant $i$ using $C_{crit}$ as,
\vspace{-10pt}
\begin{equation}
    h_i = mean(C_{crit}(Q^C_i))/c_{max} \cdot \frac{\pi}{2}, 
\end{equation}

\no where, $c_{max} = 300$ the maximum value in $C_{crit}$. Next, in each quadrant $Q^C_i$ of $C_{low}$ we modify the cost as,

\vspace{-10pt}
\begin{gather}
    C_{VA}(Q^C_i) = C_{low}(Q^C_i) \cdot \frac{clear(\kappa_i, h_i)}{max(clear(\kappa_i, h_i))}\\
    clear() = 
    \begin{cases}
    w_{s \,\text{or} \,d} \cdot (1 - \kappa_i) + \frac{2 \cdot h_i}{\pi} \,\, & \text{if} \, \Tilde{v}_i \, \text{is PV} \\
    (w_{NPV} \cdot \kappa_i + b_{NPV}) + \sin(h_i) \,\, & \text{if} \, \Tilde{v}_i \, \text{is NPV}. \label{eqn:clearing}
    \end{cases}
\end{gather}
\vspace{-10pt}

\no Here, $C_{VA}$ is a vegetation-aware cost map, $w_s, w_d, w_{NPV}$, and $b_{NPV}$ are positive constants satisfying the condition $w_d > w_s$, and $b_{NPV} > w_d + 1$. The weights $w_s$ and $w_d$ are used when $\Tilde{v}_i$ is sparse and dense grass respectively. We incorporate $sin(.)$ for NPV to ensure that the significantly tall obstacles have higher costs and differentiable cost values from short obstacles. In contrast, we consider a linearly varying cost function w.r.t. the vegetation height for PV since the resistance to the robot from pliable tall objects is correlated with their height. We note that $\sin(h_i) \ge \frac{2 \cdot h_i}{\pi}$ for $h_i \in [0, \pi/2]$.

For PV classifications, low confidence and tall vegetation lead to higher navigation costs. Intuitively, this leads to the robot preferring to navigate through high-confidence, and short pliable vegetation whenever possible. Conversely, for NPV classifications, high confidence, and tall vegetation lead to higher costs since such regions should definitely be avoided. Accounting for low-confidence classifications in the formulation helps handle misclassifications and assign costs accordingly.

\begin{prop}
The clear() function modifies $C_{low}$ such that traversability costs in PV regions are always lower than in NPV regions.  
\end{prop}

\begin{proof}
The maximum cost for PV is $w_d + 1$ (when $\kappa_i \to 0$ and $h_i \to \pi/2$), and the minimum cost for NPV is $b_{NPV}$ (when $\kappa_i \to 0$ and $h_i \to 0$). Conditions $w_s < w_d$, and $b_{NPV} > w_d + 1 \implies$ max cost of PV $<$ min cost of NPV. Therefore, in the absence of free space, the robot always navigates through PV and avoids NPV regions. Therefore, regions with PV will always be preferred for navigation.
\end{proof}

\subsection{Cautious Navigation}
We adapt DWA (section \ref{sec:dwa}) to use our vegetation-aware cost map for robot navigation. We calculate the obstacle cost (obs(.)) associated with every $(v, \omega)$ pair by projecting its predicted trajectory $traj^C(v, \omega)$ relative to the cost map over $C_{VA}$ and summing as,
\vspace{-3pt}
\begin{equation}
    obs(v, \omega) = \sum_{(row, col) \in traj^C(v, \omega)} C_{VA}(row, col).
\end{equation}
\no Next, we compute the total cost $Q(v, \omega)$ (equation \ref{eq:dwa_obj_func}). The $(v, \omega)$ that minimizes this cost is used for navigation. In some cases, the robot might have to navigate a region with high traversability cost (represented say in $Q^C_i$) which typically occurs with NPV. To imbibe cautious navigation behaviors for such scenarios, we stunt the robot's complete velocity space as $V_s = [[0, \kappa_i \cdot v_{max}], \kappa_i \cdot [-\omega_{max}, \omega_{max}]]$. 

\subsection{Recovery Behaviors}
In highly dense vegetation or in regions with low confidence classifications, the robot could still freeze or get physically entrapped. We observe that Spot's high degrees of freedom and superior dynamics allows it to recover itself from such situations if it avoids rotational motions. Therefore, we propose holonomic recovery behaviors to recover Spot from such situations. To this end, the robot periodically stores $\textit{safe}^O$ locations relative to the odom frame whenever the conditions for freezing or entrapment (section \ref{sec:adverse-phen}) are not satisfied. If the conditions are satisfied, the robot stores its current location into an $\textit{unsafe}^O$ location list and chooses a safe location such that,

\vspace{-15pt}
\begin{gather}
    unsafe^C = T^C_O \cdot unsafe^O, \\
    C_{VA}(unsafe^C) = \infty,  
\end{gather}
\vspace{-15pt}
\begin{multline}
    \textbf{p}^O_{rec} = \underset{\textbf{p}^O \in safe}{\operatorname{argmin}} \left(dist(\textbf{g}^O, \textbf{p}^O) \right), \\
    s.t \,\,  (\textbf{P}^O - \textbf{P}^O_{rob}) \in Free^O \cap Grass^O,
    \label{eqn:recover-pt}
\end{multline}
\vspace{-15pt}
\begin{equation}
    [v_x, v_y] = k_p \cdot [\textbf{p}_{rec} - \textbf{p}_{rob}].
\end{equation}

\no Here, $\textbf{p}^O_{rec}$, $\textbf{g}^O$, and $\textbf{p}^O_{rob}$ are the safe location chosen to recover to, the robot's goal and current location. The condition in equation \ref{eqn:recover-pt} ensures that the line connecting the robot and the recovery point only lies in traversable regions. Once the robot recovers, it proceeds to its goal after marking the \textit{unsafe} location with high costs in $C_{VA}$ to circumvent it. 



\section{Results and Evaluations}
We detail our method's implementation and experiments conducted on a real robot. Then, we perform ablation studies and comparisons to highlight VERN's benefits.

\begin{figure*}[t]
    \centering
    \includegraphics[width=14.75cm,height=4.75cm]{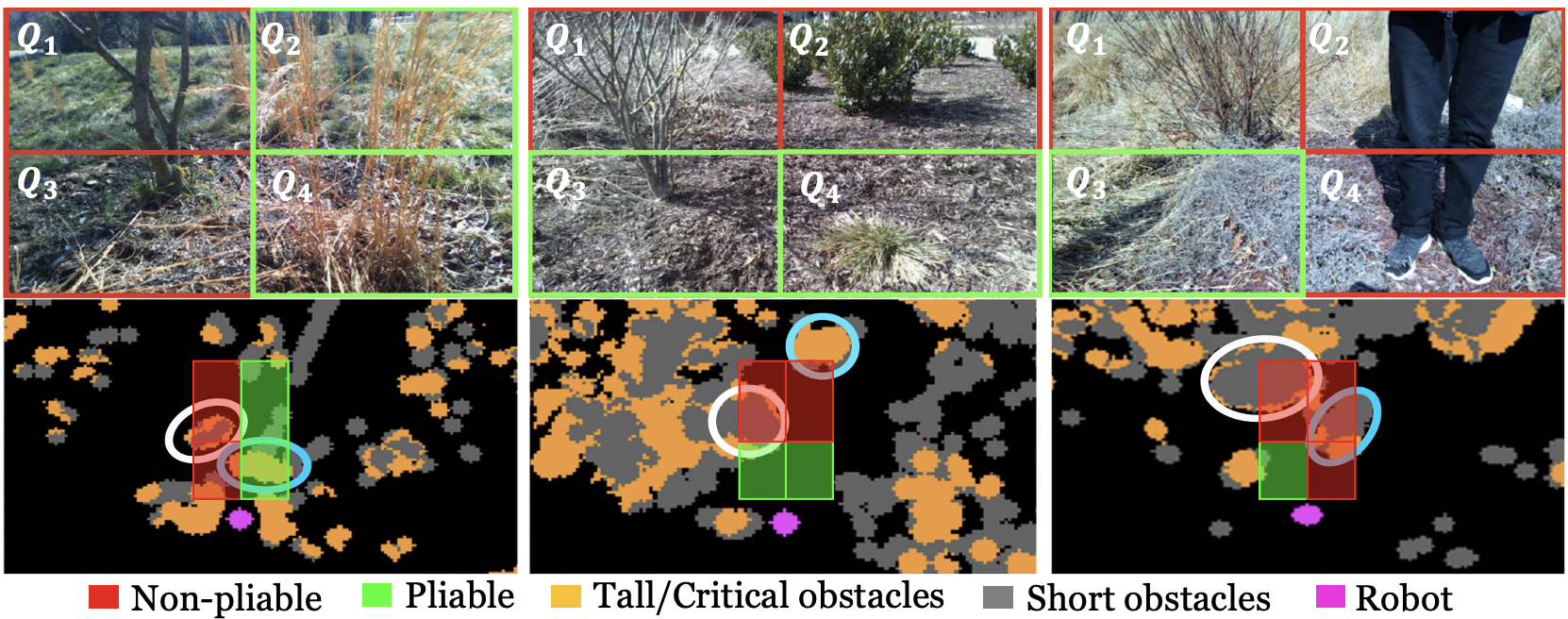}
    \caption{\small{Snapshots of the vegetation classification results on RGB images (top), and the corresponding cost maps $C_{VA}$ marked with critical obstacles (bottom) from trials in scenarios 2 (left), 3 (center), and 4 (right). A green/red rectangle in a quadrant $Q_{1,2,3,4}$ in the RGB image represents a PV/NPV classification respectively. The same quadrants are projected onto the cost maps ($Q^C_{1,2,3,4}$) along with their classifications for visualization. The cost for the vegetation within the green regions is reduced significantly using equation \ref{eqn:clearing}. The encircled regions in each cost map correspond to an obstacle in the RGB image. [\textbf{Left}]: White - Tree, Blue - Tall grass, [\textbf{Center}]: White - Tree, Blue - Bush, [\textbf{Right}]: White - Tree, Blue - Human. Our classifier accurately detects trees, and bushes as non-pliable/untraversable, and tall grass as traversable. In scenario 4, it accurately detects the human as untraversable due to its dissimilarity with all training classes and tall height.}}
    \label{fig:costmap_comparisons}
    \vspace{-10pt}
\end{figure*}

\subsection{Implementation and Dataset}
VERN's classifier is implemented using Tensorflow 2.0. It is trained in a workstation with an Intel Xeon 3.6 GHz processor and an Nvidia Titan GPU using real-world data collected from a manually teleoperated Spot robot. The robot is equipped with an Intel NUC 11 (a mini-PC with Intel i7 CPU and NVIDIA RTX 2060 GPU), a Velodyne VLP16 LiDAR, and an Intel RealSense L515 camera. The training images ($\sim600$ per class) were collected from various environments in the University of Maryland, College Park campus. Our model takes about 6 hours to train for 55 epochs. 


\begin{figure*}[t]
    \centering
    \includegraphics[width=17cm,height=4cm]{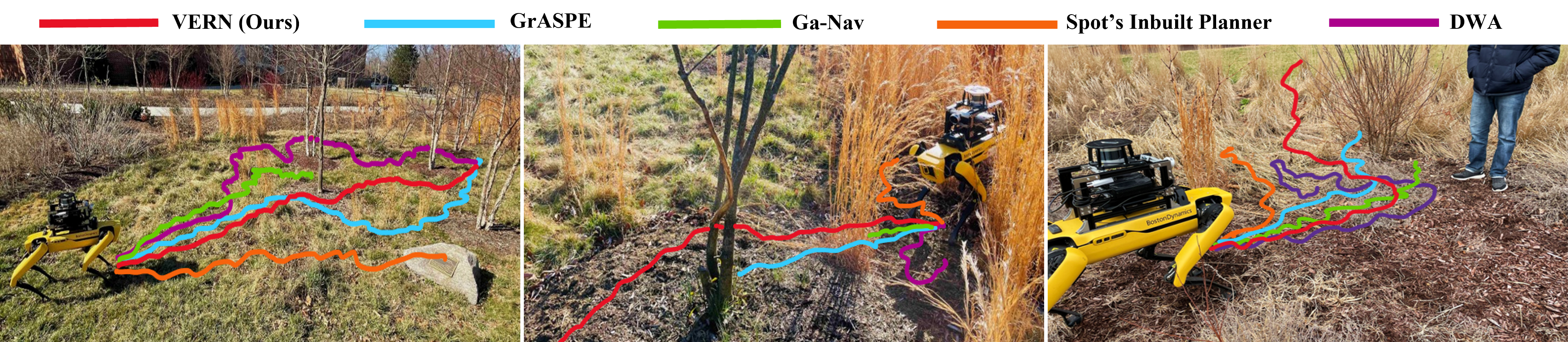}
    \caption{\small{Spot robot navigating in: \textbf{[Left]} Scenario 1, \textbf{[Center]} Scenario 2, and \textbf{[Right]} Scenario 4. We observe that VERN navigates through pliable/traversable vegetation in the absence of free space. This leads to significantly higher success rates, low freezing rates, and trajectory lengths. Other methods either freeze or take long, meandering trajectories to the goal.}}
    \label{fig:navigation_comparisons}
    \vspace{-10pt}
\end{figure*}


\subsection{Evaluations} 
We use the following evaluation metrics to compare VERN's performance against several navigation methods: 1. Boston Dynamics' in-built autonomy on Spot, 2. DWA \cite{fox1997dwa}, 3. GA-Nav \cite{ganav}, and 4. GrASPE \cite{graspe}. Spot's in-built autonomy uses its stereo cameras in four directions around the robot, estimates obstacles, and navigates to a goal. DWA is a local planner that utilizes a 2D LiDAR scan for obstacle avoidance. GA-Nav combines semantic segmentation for terrain traversability estimation with elevation maps for outdoor navigation. It is trained on publically available image datasets (RUGD \cite{RUGD2019IROS} and RELLIS-3D \cite{jiang2020rellis3d}). GrASPE is a multi-modal fusion framework that estimates perception sensor reliability to plan paths through unstructured vegetation. We further compare VERN without height estimation and recovery behaviors. The metrics we use for evaluations are,

\begin{itemize}
    \item \textbf{Success Rate} - The number of successful goal-reaching attempts (while avoiding non-pliable vegetation and collisions) over the total number of trials.

    \item \textbf{Freezing Rate} - The number of times the robot got stuck or started oscillating for more than 5 seconds while avoiding obstacles over the total number of attempts. Lower values are better.

    \item \textbf{Normalized Trajectory Length} - The ratio between the robot's trajectory length and the straight-line distance to the goal in all the successful trajectories.

    \item \textbf{False Positive Rate (FPR)} - The ratio between the number of false positive predictions (i.e., actually untraversable/non-pliable obstacles predicted as traversable) and the total number of actual negative (untraversable) obstacles encountered during a trial. We report the average over all the trials.
\end{itemize}

\no If the sum of the success and freezing rates do not equal 100, it indicates that the robot has collided in those cases. We also compare these methods qualitatively using the trajectories pursued while navigating. For perception comparisons, we quantitatively evaluate the accuracy and F-score of MobileNetv3 \cite{mobilenetv3}, EfficientNet \cite{tan2019efficientnet}, and Vision Transformer \cite{vision_transformer} when trained and evaluated on our dataset. We also show VERN's classification results and cost map clearing in Fig. \ref{fig:costmap_comparisons}.






\vspace{-2pt}
\subsection{Navigation Test Scenarios}
We compare the performance of all the navigation methods in the following real-world outdoor scenarios (see Figs. \ref{fig:cover_image} and \ref{fig:navigation_comparisons}) that differ from the training environments. The scenarios are described in increasing order of difficulty. Ten trials are conducted in each scenario for each method.

\begin{itemize}
\item \textbf{Scenario 1} - Contains sparse tall grass, and trees.

\item \textbf{Scenario 2} - Contains trees and dense tall grass in close proximity. The robot must identify grass and pass through it to reach its goal successfully.

\item \textbf{Scenario 3} - Contains trees, bushes, and dense grass in close proximity. The robot must identify grass and pass through it. See Fig. \ref{fig:cover_image} 

\item \textbf{Scenario 4} - Contains dense grass, trees, and an obstacle (human) unseen during training. 
\end{itemize}

\begin{table}[t]
\resizebox{\columnwidth}{!}{%
\begin{tabular}{ |c |c |c |c |c |c |} 
\hline
\textbf{Metrics} & \textbf{Method} & \multicolumn{1}{|p{1cm}|}{\centering \textbf{Scenario} \\ \textbf{1}} & \multicolumn{1}{|p{1cm}|}{\centering \textbf{Scenario} \\ \textbf{2}} & \multicolumn{1}{|p{1cm}|}{\centering \textbf{Scenario} \\ \textbf{3}} & \multicolumn{1}{|p{1cm}|}{\centering \textbf{Scenario} \\ \textbf{4}}\\ [0.5ex] 
\hline
\multirow{6}{*}{\rotatebox[origin=c]{0}{\makecell{\textbf{Success}\\\textbf{Rate (\%)} \\ (Higher is \\ better)}}} 
& Spot's Inbuilt Planner & 50 & 0 & 0 & 0   \\
 & DWA \cite{fox1997dwa} & 80 & 0 & 0 & 0   \\
 & GA-Nav\cite{ganav}    & 60 & 20 & 30 & 30 \\
 & GrASPE\cite{graspe}   & 80 & 60 & 50 & 40 \\
 & VERN w/o height estimation & 80 & 60 & 40 & 30 \\
 & VERN w/o recovery behavior & 100 & 60 & 40 & 40 \\
 & VERN(ours) & \textbf{100} & \textbf{90} & \textbf{70} & \textbf{70} \\
\hline

\multirow{6}{*}{\rotatebox[origin=c]{0}{\makecell{\textbf{Freezing}\\\textbf{Rate (\%)} \\ (Lower is \\ better)}}} 
& Spot's Inbuilt Planner & 30 & 100 & 100 & 90   \\
 & DWA \cite{fox1997dwa} & 20 & 100 & 100 & 100   \\
 & GA-Nav\cite{ganav}    & 40 & 80 & 70 & 70 \\
 & GrASPE\cite{graspe}   & 10 & 20 & 20 & 50 \\
 & VERN w/o height estimation & 20 & 40 & 40 & 70 \\
 & VERN w/o recovery behavior & 0 & 40 & 60 & 60 \\
 & VERN(ours) & \textbf{0} & \textbf{10} & \textbf{20} & \textbf{30} \\
\hline

\multirow{6}{*}{\rotatebox[origin=c]{0}{\makecell{\textbf{Norm.}\\\textbf{Traj.}\\\textbf{Length}\\ (Closer to 1 \\ is better)}}} 
& Spot's Inbuilt Planner & 1.23 & 0.37 & 0.24 & 0.22   \\
 & DWA \cite{fox1997dwa}  & 1.52 & 0.46 & 0.34 & 0.62   \\
 & GA-Nav\cite{ganav} & 1.31 & 1.39 & 1.32 & 1.49 \\
 & GrASPE\cite{graspe} & 1.18 & 1.09 & 1.46 & 1.37 \\
 & VERN w/o height estimation & 1.39 & 1.41 & 1.35 & 1.30 \\
 & VERN w/o recovery behavior & 1.42 & 1.48 & 1.46 & 1.39 \\
 & VERN(ours) & 1.11 & 1.19 & 1.28 & 1.23 \\
\hline

\multirow{4}{*}{\rotatebox[origin=c]{0}{\makecell{\textbf{ False}\\\textbf{Positive }\\\textbf{Rate}}}} 
 & GA-Nav\cite{ganav} & 0.33 & 0.39 & 0.30 & 0.61 \\
 & GrASPE\cite{graspe} & 0.25 & 0.31 & 0.28 & 0.44 \\
  & EfficientNet\cite{tan2019efficientnet}& 0.28 & 0.23 & 0.32& 0.35 \\
 & VERN(ours) &  \textbf{0.15} &  \textbf{0.18} &  \textbf{0.12} &  \textbf{0.21} \\
\hline

\end{tabular}
}
\caption{\small{Navigation performance of our method compared to other methods on various metrics. VERN outperforms other methods consistently in terms of success rate, freezing rate, false positive rate, and normalized trajectory length in different unstructured outdoor vegetation scenarios.}
}
\label{tab:comparison_table}
\vspace{-10pt}
\end{table}

\begin{table}[t]
\centering
\resizebox{0.75\columnwidth}{!}{
\begin{tabular}{|c|c|c|} 
\hline
\textbf{Methods}\Tstrut \Tstrut & \textbf{Accuracy} & \textbf{F-score} \\ [0.5ex] 
\hline
MobileNetv3 \cite{mobilenetv3} & \textbf{0.957} & \textbf{0.833} \\
\hline
EfficientNet \cite{tan2019efficientnet} & 0.758 & 0.632 \\
\hline
Vision Transformer (ViT) \cite{vision_transformer} & 0.689 & 0.546 \\
\hline
\end{tabular}}

\caption{ \small{\label{Tab:Results2} The accuracy and F-scores (higher values are better) of three feature-extracting backbones used to train our dataset. We observe that our MobileNetv3 has the best accuracy and F-score because of its depth-wise separable convolutions. This leads to faster and better learning.}}
\vspace{-15pt}
\end{table}

\subsection{Analysis and Discussion}
\textbf{Classification Accuracy:} We observe from Table \ref{Tab:Results2} that MobileNetv3 has the best accuracy and F-score compared to other methods. EfficientNet is difficult to train and tune especially due to longer training time requirements. Similarly, ViT requires a lot more data for training and is not compatible with the siamese network framework. 

\textbf{Navigation Comparison:} The quantitative navigation results are presented in Table \ref{tab:comparison_table}. We observe that VERN's (with height estimation used in equation \ref{eqn:clearing} and recovery behaviors) success rates are significantly higher compared to the other methods. Spot's in-built planner and DWA consider all vegetation as obstacles. This leads to Spot becoming unstable or crashing when using its in-built planner and leads to excessive freezing and oscillations when using DWA. 

GA-Nav considers all vegetation (except trees) as partially traversable because it segments most flora as grass. It also cannot differentiate vegetation due to the lack of such precisely human-annotated datasets. Additionally, in cases where vegetation like tall grass partially occludes the RGB image (see scenario 2 in Fig. \ref{fig:navigation_comparisons} [center]), GA-Nav struggles to produce segmentation results leading to the robot freezing or oscillating. In regions with tall grass, GA-Nav's elevation maps also become erroneous.

GrASPE performs the best out of all the existing methods. It is capable of passing through sparse grass and avoiding untraversable obstacles. However, in highly dense grass that is in close proximity to trees, GrASPE is unable to accurately detect and react quickly to avoid collisions with trees.

Since some existing methods do not reach the goal even once, their trajectory lengths are less than 1. The values are still reported to give a measure of their progress toward the goal. Spot's in-built planner progress the least before the robot froze or crashed. In some cases, the methods take meandering trajectories (e.g. DWA) in scenario 1 before reaching the goal. Notably, VERN deviates the least from the robot's goal and that reflects in the low trajectory length. 

\textbf{FPR}: We compare VERN's vegetation classifier's false positive rate (FPR) with GA-Nav and the EfficientNet-based classifier using manual ground truth labeling of the vegetation in the trials. We observe that GA-Nav leads to a significantly high FPR in all four scenarios primarily because GA-Nav's terrain segmentation predictions are trained on the RUGD dataset. Moreover, GA-Nav's incorrect segmentation under varying lighting conditions and occlusions increases the false positive predictions. GrASPE and the EfficientNet-based classifier are trained using the same images we used to train VERN. However, we observe that their accuracy is comparatively lower than our VERN model. This is primarily due to the fine-grained feature learning capabilities of VERN's MobileNetv3-based backbone. Additionally, GrASPE's predictions lead to erroneous results when its 3D point cloud cannot identify the geometry of the vegetation such as trees and bushes under visually cluttered instances.  

\textbf{Ablation study}: We compared VERN and its variants without using height estimation, and recovery behaviors. We observe that when the height estimation is not used for cost map clearing, the robot's success rate drops. This is mainly because height helps differentiate short and tall pliable vegetation (where the robot could freeze). Additionally, estimating the critical regions based on the height helps avoid obstacles such as humans (Scenario 4) which are not part of our classifier. 

Using recovery behaviors helps reduce freezing in the presence of dense obstacles (especially in scenarios 2, 3, and 4). Additionally, moving the robot to a safe location, and marking and remembering the unsafe region allows the robot to preemptively avoid the region in subsequent trials.

\section{Conclusions, Limitations and Future Work}

We present a novel algorithm to navigate a legged robot in densely vegetated environments with various traversability. We utilize a few-shot learning classifier trained on a few hundred quadrants of RGB images to distinguish vegetation with different pliability with minimal human annotation. This classification model is combined with estimated vegetation height (from multiple local cost maps), and classification confidence using a novel cost map clearing scheme. Using the resulting vegetation-aware cost map, we deploy a local planner with recovery behaviors to save the robot if it freezes or gets entrapped in dense vegetation.  

Our algorithm has a few limitations. Our primary navigation assumes non-holonomic robot dynamics to utilize DWA's dynamic feasibility guarantees. However, the legged robot dynamics are holonomic and further investigation is required to extend our method to relax its action constraints. We assume that the different kinds of vegetation are not intertwined with one another. This may not be the case in highly forested environments. VERN could lead to collisions in the presence of thin obstacles such as branches that are not detected in the cost maps. In the future, we would like to augment our current method with proprioceptive sensing for navigating more complex terrains with occluded surfaces.

\bibliographystyle{IEEEtran}
\bibliography{References}

\end{document}